\documentclass[10pt, a4paper]{article}

\usepackage{lrec-coling2024} 

\usepackage{times}
\usepackage{latexsym}

\usepackage[T1]{fontenc}
\usepackage[utf8]{inputenc}
\usepackage{microtype}
\usepackage{inconsolata}
\usepackage{todonotes}
\usepackage{amsmath}
\usepackage{amssymb}
\usepackage{booktabs}
\usepackage{lipsum}
\usepackage{enumitem}
\usepackage{xspace}
\usepackage{listings}
\usepackage{soul} 
\usepackage{caption}
\usepackage{transparent} 
\usepackage{tcolorbox} 
\usepackage{mdframed} 
\usepackage{csquotes}
\usepackage{multirow}
\usepackage{colortbl}
\usepackage{pifont}
\usepackage{tipa}
\usepackage{subfig}
\usepackage{graphicx}
\usepackage{float}
\usepackage{hyphenat}

\definecolor{TodoColor}{rgb}{1,0.7,0.6}

\sethlcolor{TodoColor}

\definecolor{DocumentLinkColor}{rgb}{0.4,0.6,0.3}
\definecolor{ethblue}{rgb}{0,0.1,0.4}

\newmdenv[
  linecolor=black,
  linewidth=1.2pt,
  topline=false,
  bottomline=false,
  rightline=false,
  innertopmargin=0mm,
  innerbottommargin=-0.5mm,
  skipabove=1.1\topsep,
  skipbelow=0.5\topsep,
]{quotebox}

\title{Reference-less Analysis of Context Specificity in Translation\\with Personalised Language Models}

\newcommand{\AfA}{\textsuperscript{\begin{minipage}[c]{2.5mm}
\includegraphics[width=\linewidth]{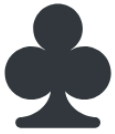}
\end{minipage}}}
\newcommand{\AfB}{\textsuperscript{\begin{minipage}[c]{2.5mm}
\includegraphics[width=\linewidth]{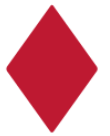}
\end{minipage}}}
\newcommand{\bustspeaker}{\begin{minipage}[c]{4.5mm}
\includegraphics[width=\linewidth]{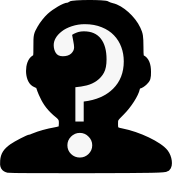}
\end{minipage}\hspace{1mm}}
\newcommand{\annedspeaker}{\begin{minipage}[c]{4.5mm}
\includegraphics[width=\linewidth]{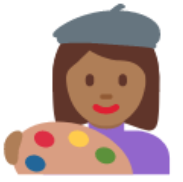}
\end{minipage}\hspace{1mm}}
\newcommand{\cmark}{\textcolor{OliveGreen}{\ding{51}}}
\newcommand{\xmark}{\textcolor{BrickRed}{\ding{55}}}

\name{\AfA Sebastian Vincent,
   	\AfB \textbf{Alice Dowek},
    \AfB \textbf{Rowanne Sumner},
   	\AfB \textbf{Charlotte Blundell}, \\
   	\AfB \textbf{\large Emily Preston},
   	\AfB\textbf{\large Chris Bayliss},
   	\AfB \textbf{\large Chris Oakley},
   	\AfA \textbf{\large Carolina Scarton}}
   	\address{\AfA Department of Computer Science, University of Sheffield, UK\\
   	\AfB ZOO Digital Group PLC, UK\\}

\abstract{
Sensitising language models (LMs) to external context helps them to more effectively capture the speaking patterns of individuals with specific characteristics or in particular environments. This work investigates to what extent rich character and film annotations can be leveraged to personalise LMs in a scalable manner. We then explore the use of such models in evaluating context specificity in machine translation. We build LMs which leverage rich contextual information to reduce perplexity by up to 6.5\% compared to a non-contextual model, and generalise well to a scenario with no speaker-specific data, relying on combinations of demographic characteristics expressed via metadata. Our findings are consistent across two corpora, one of which (\textsc{Cornell-rich}) is also a contribution of this paper. We then use our personalised LMs to measure the co-occurrence of extra-textual context and translation hypotheses in a machine translation setting. Our results suggest that the degree to which professional translations in our domain are context-specific can be preserved to a better extent by a contextual machine translation model than a non-contextual model, which is also reflected in the contextual model's superior reference-based scores.\\ \newline 
}

\begin{document}
\maketitleabstract

\section{Introduction}
Contrary to studies in sociolinguistics \citep{milburn-2004-speech}, conventional approaches to generation tasks build models in a one-size-fits-all fashion, and most often for a particular language and domain,
disregarding the context of the processed text. This leads to assuming the most likely scenario as context, sometimes resulting in harmful predictions \citep[e.g. the \enquote{masculine default} in][]{Schiebinger2014}. \textbf{Personalisation} -- adapting model predictions to the unique dialogues of individuals -- offers clear benefits in generation tasks \citep{flek-2020-returning}, where context information helps to disambiguate the input text, aiding correct interpretation and minimising sample bias in training data \citep{dudy-etal-2021-refocusing}.

Incorporating demographic factors has been shown to improve the performance of models developed for NLP tasks such as classification \citep{hovy-2015-demographic} and translation \citep{vincent-etal-2022-controlling-extra}. Their influence can manifest in two main dimensions: \textbf{grammatical} and \textbf{behavioural}. Grammatical agreement is well-defined and pertains to morphosyntactic structures, whereas behavioural is more fluid, relating to how language is utilised across different demographics or contexts \citep{vincent2023thesis}. For instance, the speaker's gender may dictate morphological endings in self-referential verbs in certain languages, exemplifying grammatical agreement. Conversely, behavioural agreement is more nuanced: \enquote{They're done!} has a different meaning when said by a baker about a batch of cookies than by a frenzied king about his treacherous subjects. Contextual language generation research \citep[e.g.][]{voita-etal-2019-good} mostly emphasises grammatical agreement, whereas in practice both types of agreement are required in the language generation process and a robust framework should encompass them both.

This work explores personalisation of language models (LMs) and machine translation (MT) systems for \textbf{speakers} and \textbf{productions} in TV series and film, focusing on leveraging the similarities in their metadata, an approach termed \textbf{context-based} personalisation. The language usage within our domain exhibits significant variability. For instance, TV writers often create characters that emulate the speech patterns of particular social groups they represent. Similarly, productions from specific decades, countries, or genres capture unique discourse nuances. We demonstrate how speaker and production metadata can inform the development of LMs which excel at capturing the language style of a given speaker or production compared to generic models. We then apply these LMs to measure the \textbf{context specificity} of hypothesis tokens used in professional and machine translations within the TV series domain. This evaluation is done in a reference-less manner, potentially offering a future evaluation tool for context-aware MT free of reference bias.

Our domain presents an additional challenge: ensuring model robustness in scenarios where no prior dialogue samples exist for certain speakers or productions. This setting, where there is insufficient content to characterise the subjects of a given system, is commonly known as the cold start problem \citep[e.g.][]{schein-etal-2002-cold, huang-etal-2014-enriching}. Models adapted solely on past dialogue fall short in this case, and we argue that a context-based approach is more effective, mimicking the benefits of personalisation by estimating token distributions for similar character/production profiles.

We collect a rich set of metadata annotations for two corpora based on identifiable information (i.e. character names and titles). This allows us to perform experiments on up to 14 unique metadata variables at once, to our knowledge the richest set of metadata information for personalisation. In contrast, metadata-based approaches to personalisation reported in previous work in different domains were small-scale, leveraging a few simple and mostly categorical variables \citep{huang-etal-2014-enriching,
lynn-etal-2017-human,
king-cook-2020-evaluating,
welch-etal-2020-compositional,
guo-etal-2021-aerospace}.

Our work is presented in two parts: firstly, we consider whether rich character profiles can be used to model the characters' speaking styles, including for characters which did not appear in the training data, by learning from data for characters with similar profiles (\S \ref{lrec:lmcue}). Then, we explore how such personalised LMs can be used to estimate the context specificity (or extent of personalisation) of professional and machine translations (\S \ref{lrec:translations}). Additionally, we contribute \textsc{Cornell-rich} (\S \ref{lrec:cornell-rich}), a corpus of rich character and film annotations for the Cornell Movie Dialogue Corpus \citep{Danescu-Niculescu-Mizil2011} (\textsc{Cornell}) and \textsc{sMRR}, an evaluation metric for personalised LMs.\footnote{\url{https://github.com/st-vincent1/cornell_rich}} This paper also presents the related work (\S \ref{lrec:related}), and conclusions (\S \ref{lrec:conclusions}).

\section{Related Work} \label{lrec:related}
\paragraph{Personalisation in NLP}
Personalisation in NLP can generally be split into three groups: with respect to how much data is available for a speaker: 
\textit{full supervision}, where there is sufficient training data to fine-tune a model for a particular speaker, 
\textit{few-shot}, where some supervised data exists but not in quantities sufficient for supervised training, and 
\textit{zero-shot}, where no samples of text exist for the speaker used in evaluation. 
Full supervision is usually facilitated through some form of a \textit{user embedding} or \textit{tagging} approach \citep[e.g.][]{sennrich-etal-2016-controlling,Keskar2019ctrl,mireshghallah-etal-2022-useridentifier}. Among few-shot approaches, \citet{king-cook-2020-evaluating} examine several personalisation methods for language modelling of blog posts with sample adaptation data for new users, including fine-tuning, interpolation (averaging the fine-tuned speaker model with a general model) and demographic-based adaptation (fine-tuning on text from users with the same age and gender). \citet{welch-etal-2022-leveraging} leverage models built for \enquote{anchor users} (who boast a large history of posting) to devise models for new users (with a small number of posts), focusing on the similarity between samples of users' posts. Finally, zero-shot approaches typically leverage background data available for the new speakers, e.g. their metadata. \citet{huang-etal-2014-enriching} rely on the social network of a user to model their language; \citet{lynn-etal-2017-human} use age, gender, and personality traits to improve user modelling in multiple NLP tasks; \citet{zeng-etal-2019-automatic} leverage user profiles to improve comment generation on a social media corpus. Demographic-based adaptation was found inferior to interpolation and priming in the few-shot scenario by \citet{king-cook-2020-evaluating}, but their study used only two factors, age and gender. Our work is positioned in the zero-shot category as we rely on rich metadata annotations to model the dialogue of individual screen characters appearing in particular productions. Unlike \citet{king-cook-2020-evaluating}, we leverage textual (real-valued) metadata annotations, which in personalisation are preferable to categorical values \citep{lynn-etal-2017-human}, and a significantly higher count of them (up to 14). Importantly, we also leverage our personalised LMs to quantify context specificity in translation.

A few studies have explored the idea that context in NLP can be summarised with pre-trained models. \citet{novotney-etal-2022-cue} introduce the notion of \textsc{cue} (contextual universal embeddings), representing individual context variables as pre-trained sentence embedding vectors. They demonstrate that including article metadata in the form of \textsc{cue} into language modelling improves perplexity when applied to news articles. \citet{vincent-etal-2023-mtcue} explore this idea further, applying it to machine translation of dialogue and showing that pre-training on film metadata helps zero- and few-shot performance in some contextual MT tasks. In this paper, we leverage context in the same way as \citet{vincent-etal-2023-mtcue}, but focus on contextual language modelling, and specifically on personalisation for individual characters and films. We also explore a practical application of such personalised LMs in evaluation of contextual MT (\S \ref{lrec:translations}), and contribute an evaluation metric for personalised LMs (\S \ref{lrec:lm-exp}).

\paragraph{Evaluation of Contextual Machine Translation}
Traditional measures of MT quality are based on sentence-level matching to references, and offer little insight into performance at maintaining or introducing context-specific features of the source text. Alternative evaluation methods of contextual MT have been explored to address this. When contextual phenomena are directly observable and necessitate grammatical agreement (e.g. in formality transfer or document-level translation), evaluation usually involves parsing tools \citep{sennrich-etal-2016-controlling,vincent-etal-2022-controlling} or contrastive evaluation on bespoke test suites \citep{bawden-etal-2018-evaluating,muller-etal-2018-large,voita-etal-2019-good,lopes-etal-2020-document}. However, the creation of such tools and test sets is expensive, and as argued in \citet{post2023escaping}, strong performance at contrastive evaluation does not necessarily entail the ability to generate contextual translations in practice. Evaluation of behavioural agreement (e.g. preserving individual style of a character or production), has mostly been limited to classification systems \cite[e.g.][]{Michel2018} which attribute the input text as belonging to one of a list of speakers. However, such systems depend on sufficient quantities of training data for each considered speaker, which is usually not readily available.

\paragraph{Language Models in Machine Translation}
LMs have been utilised in machine translation as means of improving fluency \citep{stahlberg-etal-2018-simple}, boosting document-level performance \citep[e.g.][]{sugiyama-yoshinaga-2021-context} or evaluation \citep{edunov-etal-2020-evaluation}. In contrast, we build a tandem of LMs from the same training data as the NMT models to measure to what extent the NMT generations are context-specific. Our metric is pointwise mutual information (PMI), similarly to \citet{sugiyama-yoshinaga-2021-context} who used PMI between document context and the target utterance to boost document-level performance of MT.

\section{Building a Personalised LM} \label{lrec:lmcue}
The first stage of our work delves into building a personalised language model for dialogue associated with rich contextual annotations. We create two metadata-rich datasets (\S \ref{lrec:lm-datasets}) and train a contextual language model to capture the distribution of the tokens in the dialogue given a set of contextual variables (\S \ref{lrec:lm-exp}). This section addresses two research questions (RQs): 
\begin{enumerate}[rightmargin=1em,itemsep=.1em]
  \setlength{\itemindent}{2em}
    \item [\textit{\textbf{RQA}}] \textit{How can rich character profiles be used to model the characters' speaking styles?}
    \item [\textit{\textbf{RQB}}] \textit{How can a LM be personalised for a specific character solely by learning from data for characters with similar profiles?}
\end{enumerate}

\subsection{Datasets} \label{lrec:lm-datasets}
In the absence of similar datasets, we use a combination of manual and automatic annotation to create two English-language corpora: \textsc{ZOO} and \textsc{Cornell-rich}. The domain of both corpora is TV series and film dialogue respectively, and samples within each corpus consist of: an utterance in English and a set of up to 14 textual metadata annotations for the speaking character (age bracket, country of origin, description, gender, profession, religion and characteristic quote) and for the production (country, genre, PG rating, plot description, writers, year). All metadata contained in the corpora was collected by the authors of this paper manually or automatically. Metadata within \textsc{Cornell-rich} (the corpus we make publicly available) was collected from readily available materials with a permissive license. Below we summarise the descriptions for each corpus.

\paragraph{\textsc{ZOO}} \label{lrec:ZOO}
The \textsc{ZOO} corpus is a private in-house collection of subtitles for nine anglophone TV series. 
It totals 157K dialogue lines and annotations for 159 speakers of 101K lines. It is divided into traditional \texttt{test}, \texttt{valid} and \texttt{train} sets, but features an additional test set of metadata and dialogue from 11 held-out speakers who do not appear in the remaining sets. Quantitative details are reported in \autoref{tab:data-quants} (rows 1-5).
The corpus was created from production-ready subtitle files from which dialogue with character and TV series attributions was extracted. This data was subsequently annotated with production metadata (automatically, via the OMDb API\footnote{\url{https://www.omdbapi.com/}}) and character metadata for the most frequently speaking characters. The annotation process is detailed in \autoref{app:data-collection}. 
In total, 87.1\% of productions and 50.3\% of speakers received annotations.

\begin{figure}[h!]
	\centering
	\resizebox{\linewidth}{!}{\includegraphics{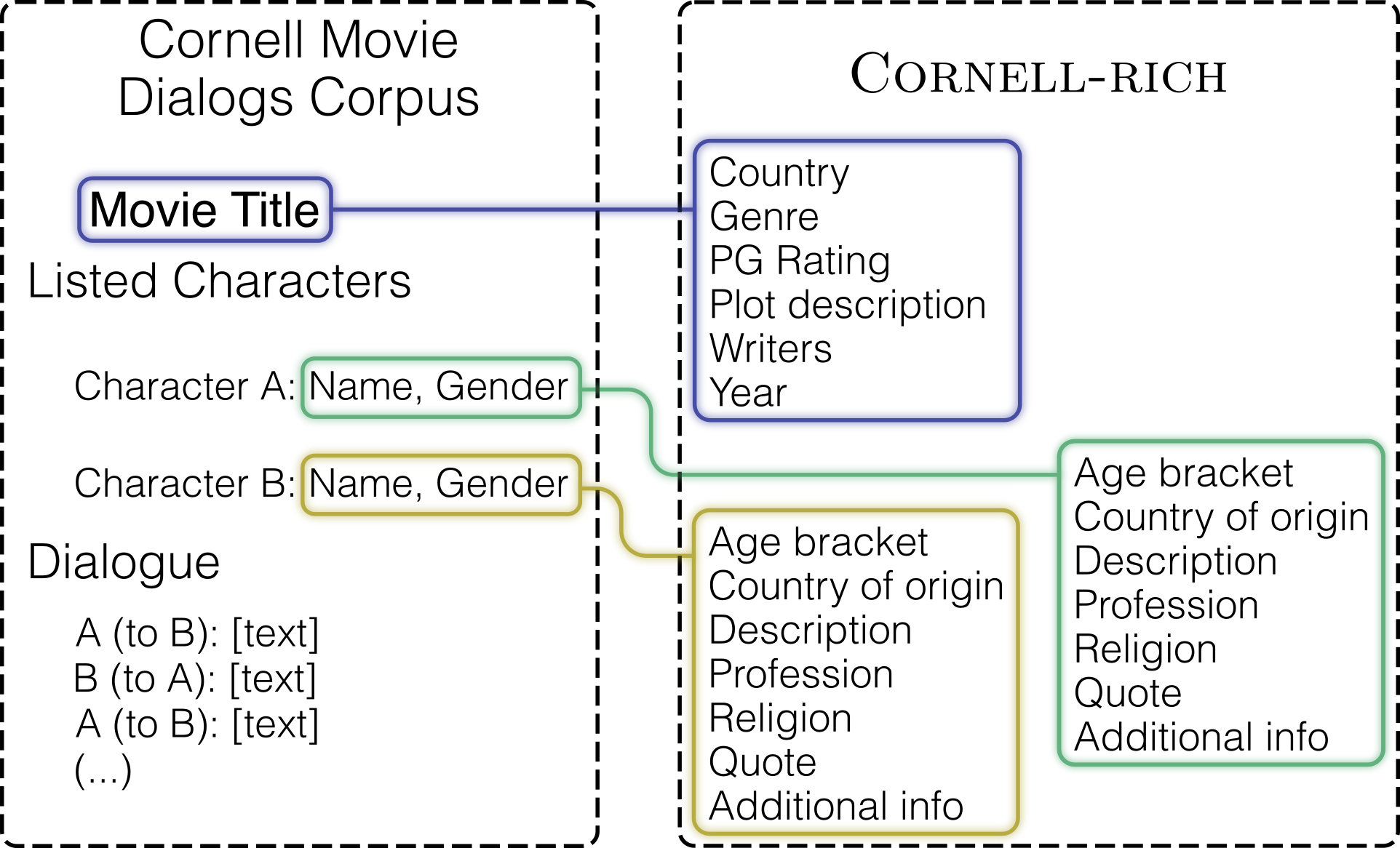}}
    \caption{\textsc{Cornell-rich} overview.}
	\label{lrec:cornells}
\end{figure}

\paragraph{\textsc{Cornell-rich}} \label{lrec:cornell-rich}
Much like \textsc{ZOO}, \textsc{Cornell-rich} is a dataset of rich character and production annotations, albeit for film dialogue extracted from scripts. It includes 14 distinct metadata variables captured as text. The collected annotations can be linked to the entries of \textsc{Cornell} \citep{Danescu-Niculescu-Mizil2011}, a corpus of exchanges from a set of film scripts, with character dialogue attributions (\autoref{lrec:cornells} illustrates how \textsc{Cornell-rich} enriches the original corpus). Both dialogue data and annotations are in English. We provide details about the collected metadata, together with examples, in \autoref{app:cornell-details}.
The corpus is split into \texttt{train}, \texttt{valid}, \texttt{test} and \texttt{test\_unseen} in the same way as \textsc{ZOO}, and \texttt{test\_unseen} features 30 held-out speakers (see
\autoref{tab:data-quants}: rows 6-10).

\paragraph{\textsc{OpenSubtitles}} \label{lrec:opensubs}
We use the OpenSubtitles18\footnote{Based on \url{https://opensubtitles.org/}} corpus \citep{lison-etal-2018-opensubtitles2018} (\textsc{OpenSubtitles}) to pre-train the LMs. It is a a large collection of subtitles with timestamps that facilitate the extraction of document-level information. Focusing on past context with no loss of generality, we extract up to 3 past sentences based on the timestamps (\autoref{tab:data-quants}, rows 11-12). Roughly 68\% samples contain at least one past sentence. 

\paragraph{Preprocessing} \label{lrec:lm-preprocessing}
Since both \textsc{ZOO} and \textsc{Cornell} are of high quality as is, our preprocessing only involves normalising punctuation, removing tokenisation using the \texttt{sacremoses} package\footnote{\url{https://pypi.org/project/mosestokenizer/}}, fixing leftover punctuation issues (e.g. ensuring all multi-dots use three dots) and removing \textsc{HTML} tags. We also preprocess all (original and added) annotations so that: (i) all empty fields are expressed as an empty string; (ii) there are no multiple expressions of the same discrete type (e.g. \textit{m} and \textit{M} to denote masculine gender); (iii) all attributes are expressed in unambiguous natural language (e.g. a PG rating of \enquote{R} is rewritten as \enquote{PG Rating: R}). \textsc{OpenSubtitles} is preprocessed 
following \citet{vincent-etal-2023-mtcue}. 
For subword tokenisation, we use SentencePiece to train a BPE model of 8K tokens on the \texttt{train} split of \textsc{Cornell-rich}; it is then used to tokenise all datasets.

\begin{table}[h!]
\centering
\resizebox{\linewidth}{!}{
\begin{tabular}{@{}crccc}
\toprule
& \multicolumn{1}{l}{} & \multicolumn{3}{c}{Total number of}	\\
Row & Dataset \& split & segments & tokens & metadata types \\
\midrule
(1) & \textsc{ZOO} & & & \\
(2) & \texttt{train}   	& $140.4$K & $1.1$M  & \multirow{4}{*}{$13$} \\
(3) & \texttt{valid}   	& $4$K 	& $31.3$K &  \\
(4) & \texttt{test} & $6$K 	& $47.1$K &  \\
(5) & \texttt{test\_unseen} & $6.7$K   & $51.5$K &  \\ \midrule
(6) & \textsc{Cornell-rich} & & &  \\
(7) & \texttt{train} & $289.0$K & $3.1$M & \multirow{4}{*}{$14$}\\
(8) & \texttt{valid} & $5$K & $51.2$K & \\
(9) & \texttt{test} & $5$K & $54.4$K & \\
(10) & \texttt{test\_unseen} & $5.2$K & $54.6$K & \\ \midrule
(11) & \textsc{OpenSubtitles} & & &  \\
(12) & \texttt{train}   	& $14.7$M   & $109.6$M & $3$* \\

\bottomrule
\end{tabular}}
\caption[Quantities of segments, tokens and unique metadata in the \textsc{OpenSubtitles}, \textsc{Cornell-rich} and \textsc{ZOO} datasets.]{Quantities of segments, tokens (pre-tokenisation) and unique metadata (speaker and production) in datasets. *\textsc{OpenSubtitles} uses three past sentences as proxy metadata.}
\label{tab:data-quants}
\end{table}

\subsection{Experimental Setup} \label{lrec:lm-exp}
\paragraph{\textsc{LMCue} Architecture} Our selected LM architecture is adapted from the \textsc{MTCue} model \citep{vincent-etal-2023-mtcue}, which is a Transformer-based multi-encoder contextual machine translation system. \textsc{MTCue} processes the source text with a source encoder and the context information with an additional context encoder. We convert \textsc{MTCue} to a LM by removing the source encoder, resulting in a conditional encoder-decoder LM where context is treated as the input to the encoder. The sequence of context information is converted to a sequence of equal-sized vectors with a sentence embedding model (\textsc{MiniLM-v2}). This approach has the advantage of treating both discrete and continuous (text) inputs in the same way, potentially utilising the semantic information of the discrete labels, as well as allowing longer spans of context as input without issues of long-range dependencies. The target sequences are contextualised via standard encoder-decoder attention which maps queries (target) to keys and values (context). 
We select this approach as \textsc{MTCue} can process large sets of contextual information and has the potential to scale well to few- and zero-shot scenarios, which in our case are explored when we consider test sets with completely new speakers. For simplicity, hereinafter we refer to this architecture as \textbf{\textsc{LMCue}}.

\begin{figure}[h!]
	\centering
	\resizebox{\linewidth}{!}{
	\includegraphics{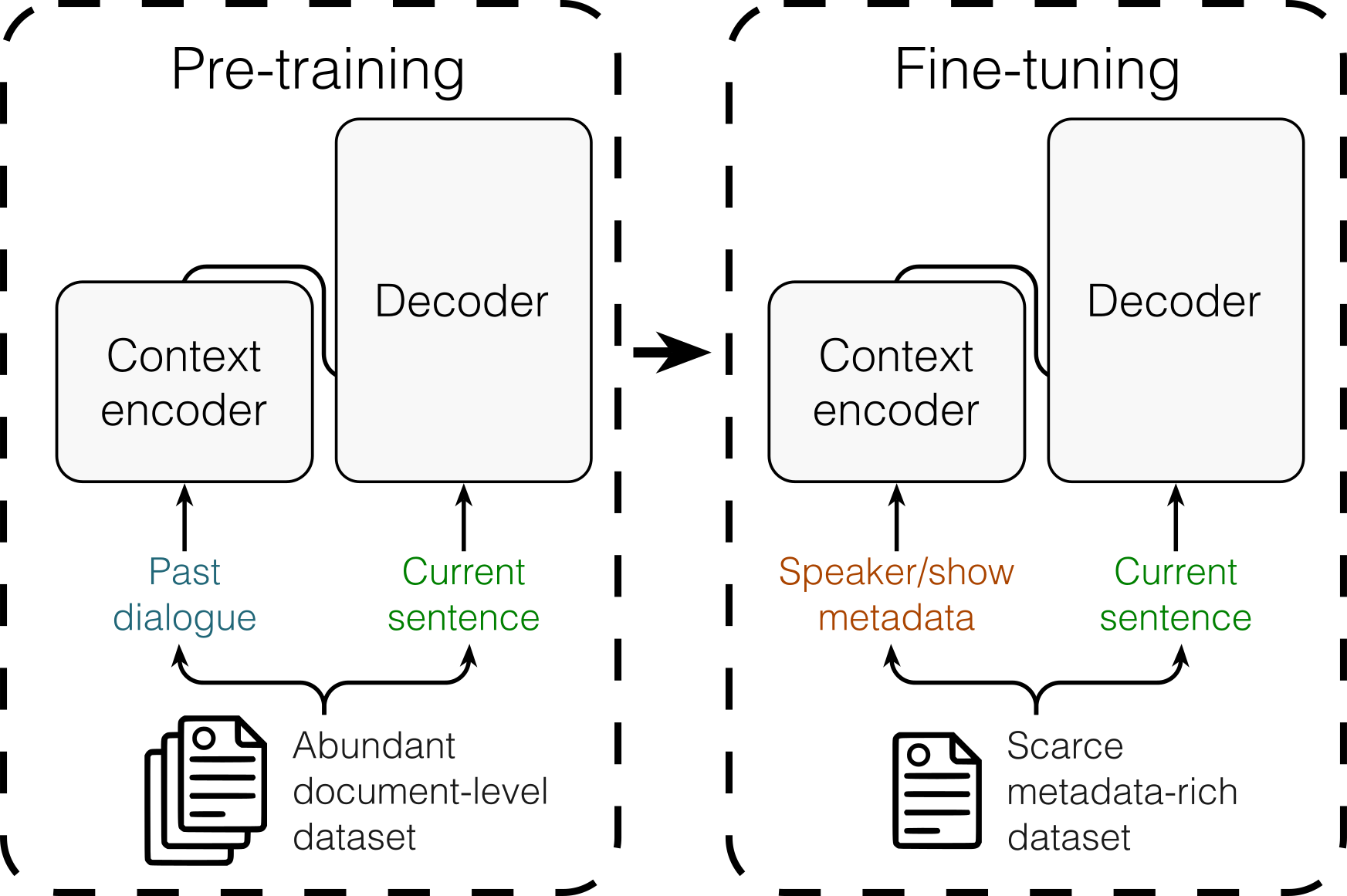}}
	\caption{An illustration of the pre-training and fine-tuning regimens used in the experiments.}
	\label{lrec:pt-ft}
\end{figure}

\paragraph{Pre-training} \label{lrec:lm-pretraining}
Preliminary experiments showed that training \textsc{LMCue} from scratch on \textsc{Cornell-rich} lead to results inferior to a non-contextual LM trained on the same data (see \autoref{app:pretraining}). 
We, therefore, experimented with pre-training the model first. Since a larger corpus of dialogue with character metadata is unavailable, we used a corpus with document-level information and treated the \textbf{past dialogue} for any sentence as the \textbf{metadata context}. We hypothesised that at a larger scale, the effect of metadata embeddings on text generation will be similar to the effect of embeddings of past dialogue (\autoref{lrec:pt-ft}), meaning the pre-training procedure allows the model to learn dependencies between the context and the text. An advantage of this approach is that metadata-rich corpora are likely to be too small to train a model from scratch, but document-level information is abundant. In our case, pre-training on past dialogue proved successful; consequently, all models considered in our experiments are pre-trained. For this purpose, we use the \textsc{OpenSubtitles} corpus (\S \ref{lrec:opensubs}).

\paragraph{Baselines and Implementation} \label{lrec:lm-baselines}
We consider three baselines: a non-contextual LM (\textsc{Base-LM}), a speaker-wise fine-tuning baseline (\textsc{SpFineTuning}) and a linear interpolation method (\textsc{Lerp}) which ensembles \textsc{SpFineTuning} with the general model \textsc{Base-LM} at test time \citep[baselines are modelled after][]{king-cook-2020-evaluating}.

We implement \textsc{LMCue} by modifying the code provided by \citep{vincent-etal-2023-mtcue}. The model has 159M parameters and comprises a context encoder (38M) and a decoder (121M). 25\% of the decoder's parameters are used by the encoder-decoder attention; a non-contextual decoder of this shape would have 91M parameters. To make the comparison fair, \textsc{Base-LM} matches the total number of parameters in \textsc{LMCue} (159M) and is therefore wider than the decoder in \textsc{LMCue} (\autoref{lrec:model-details}). This strong baseline removes the possibility that the model improves simply because of a higher parameter count\footnote{Results from using the smaller baseline LM (91M params) were consistently inferior to 159M by up to 0.75 perplexity.}. All other baselines (\textsc{SpFineTune}, \textsc{Lerp}) share the architecture and size of \textsc{Base-LM}.

\begin{table}[ht]
\centering
\resizebox{\linewidth}{!}{
\begin{tabular}{clccccc}
\toprule
 & & Params & $d_{model}$ & $n_{layers}$ & $h$ & FFN dim. \\
 \midrule
(1) & \textsc{LMCue} (Enc.) & $38$M & $512$ & $6$ & $8$ & $2048$ \\
(2) & \textsc{LMCue} (Dec.) & $121$M & $768$ & $12$ & $12$ & $3072$ \\
(3) & \textsc{LMCue} (total) & $159$M & $-$ & $-$ & $-$ & $-$ \\ \midrule
(4) & \textsc{Base-LM} & $159$M & $1024$ & $12$ & $16$ & $4096$ \\ \bottomrule
\end{tabular}}
\caption{Model details for \textsc{LMCue} and \textsc{Base-LM}.}
\label{lrec:model-details}
\end{table}

The \textsc{LMCue} models are pre-trained on \textsc{OpenSubtitles} (using past dialogue as context). \textsc{Base-LM} is pre-trained on the text part of the same corpus, one sentence at a time. Training details (including the hyperparameter search) are provided in \autoref{app:lmcue-training}.

\begin{figure}[h!]
	\centering
	\resizebox{\linewidth}{!}{
	\includegraphics{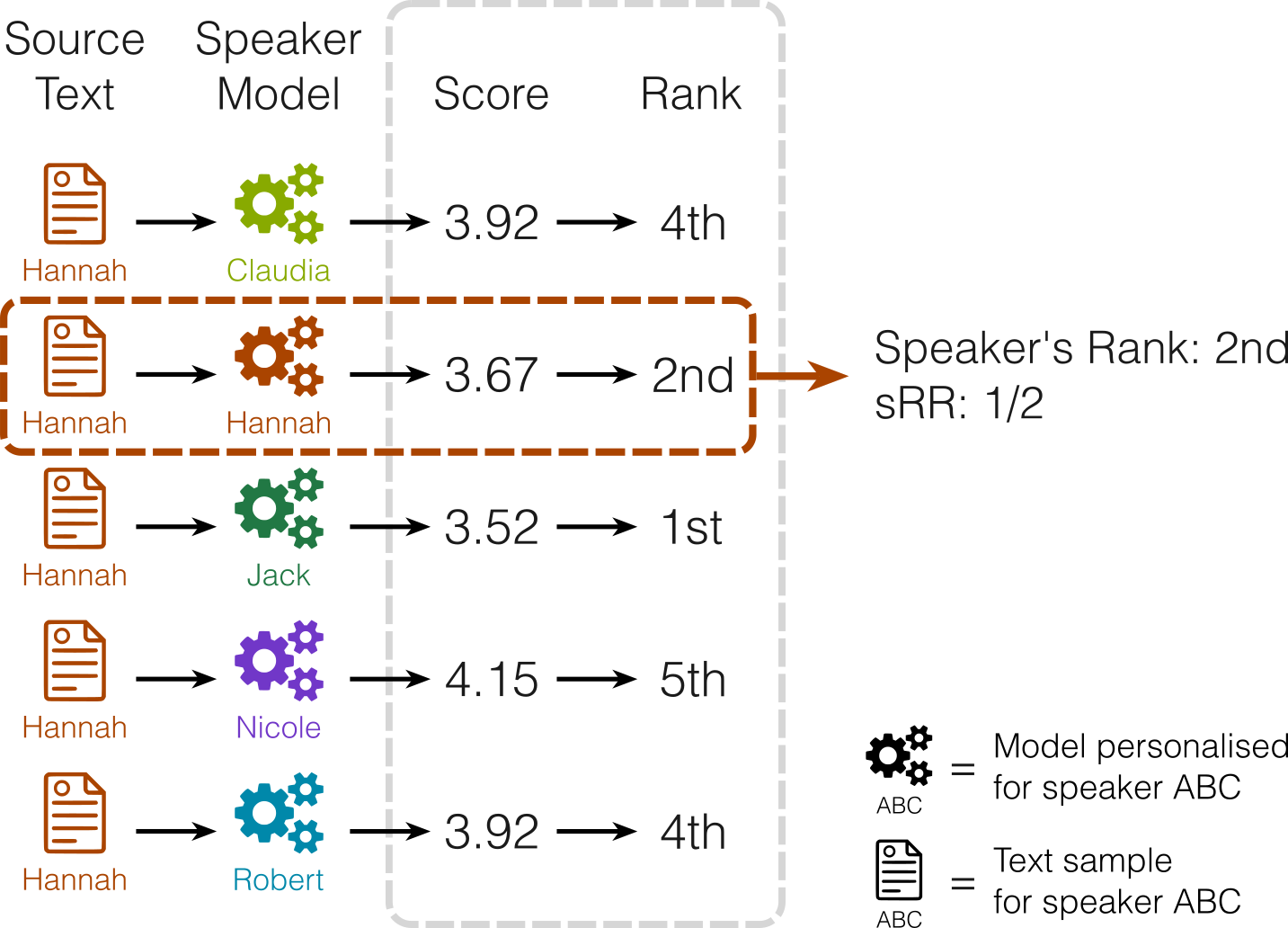}}
	\caption{\textsc{sRR} illustrated for speaker Hannah.}
	\label{lrec:smrr}
\end{figure}

\paragraph{Evaluation} \label{lrec:eval}
For evaluation, we use perplexity (\textsc{ppl}) as well as \textsc{sMRR}, which we define as follows: let $M_j$ be a model personalised for a speaker $s_j$ and $U_i$ be a set of utterances by a speaker $s_i$. We calculate speaker reciprocal rank \textsc{sRR} for any speaker $k$ by scoring the $U_k$ with $M_1,...,M_n$ (expressed with log likelihood), then ranking the models best to worst by this score\footnote{Ties are resolved pessimistically.} and taking the reciprocal rank ($1/rank$) of $M_k$, the model for speaker $k$ (see \autoref{lrec:smrr}). \textsc{sMRR} is \textsc{sRR} averaged for all speakers; $1/\textsc{sMRR}$ is the average rank of the correct speaker model. Intuitively, this metric captures the strength of the association between dialogue and the speaker model: \textsc{sMRR} of 1.0 indicates that for any speaker $j$, the model $M_j$ produces the best score for $U_j$.

Unless otherwise specified, all results are calculated from five runs with different random initialisation, and the reported value is the \textbf{mean} result. We \textbf{highlight} the best overall result. Unless another result is \underline{underlined}, it is significantly worse (indicating a less effective model) than the best result in bold, with a confidence interval of 95\% (computed with a one-tailed t-test, $t(4)=1.65$, $p=0.05$).

\subsection{Results} \label{lrec:lm-results}
This section presents the results of training \textsc{LMCue} on the \textsc{Cornell-rich} and \textsc{ZOO} corpora. 
We use $\mathcal{S}$ and $\mathcal{P}$ to denote that $\mathcal{S}$peaker or $\mathcal{P}$roduction metadata was used in training (or both, i.e. $\mathcal{S}+\mathcal{P}$).

\paragraph{Are speaker profiles helpful?} \label{lrec:rq1}
We examine whether including speaker profiles as a supplementary input in language modelling can result in significant quantitative improvements. For this, we train models on the \texttt{train} splits and evaluate on the \texttt{test} splits of both corpora, with overlapping speakers (= unique combinations of speaker profiles) between them. As demonstrated in \autoref{tab:rq1-results}, context-based personalisation with \textsc{LMCue} results in substantial reductions in perplexity compared to the best baseline, with a decrease of 5.4\% for \textsc{Cornell-rich} and 6.5\% for \textsc{ZOO}, respectively.

\begin{table}[h!]
\centering
\resizebox{\linewidth}{!}{
\begin{tabular}{r|cc|cc}
\toprule
 & \multicolumn{2}{c|}{\textsc{Cornell-rich}} & \multicolumn{2}{c}{\textsc{ZOO}} \\
 & \texttt{valid} & \texttt{test} & \texttt{valid} & \texttt{test} \\
 \midrule
\textit{\textbf{baselines}} &  &  &  & \\
\textsc{Base-LM} & $22.35$ & $23.38$ & $18.42$ & $18.41$ \\
\midrule
\textit{\textbf{proposed}}  &  &  &  & \\
\textsc{LMCue} ($\mathcal{S}$) & $21.37$ & $22.37$ & $17.52$ & $17.55$ \\
\textsc{LMCue} ($\mathcal{P}$) & $\mathbf{21.07}$ & $\mathbf{22.04}$ & $17.18$ & $17.29$ \\
\textsc{LMCue} ($\mathcal{S}+\mathcal{P}$) & $21.14$ & $22.13$ & $\mathbf{17.13}$ & $\mathbf{17.21}$ \\
\bottomrule
\end{tabular}}
\caption{Perplexity$^\downarrow$ on different validation and testing sets for the two corpora.}
\label{tab:rq1-results}
\end{table}

\paragraph{Is speaker-based adaptation better than direct fine-tuning?}

\begin{table*}[h!]
\centering
\resizebox{\linewidth}{!}{
\begin{tabular}{@{}rlp{1.2cm}p{3cm}llp{8cm}@{}}
\toprule
\textit{\textbf{ID}} & \textit{\textbf{\#Lines}} & \textit{\textbf{Age}} & \textit{\textbf{Profession}} & \textit{\textbf{Country}} & \textit{\textbf{Genre}} & \textit{\textbf{Characteristic quote}} \\ \midrule
\textbf{sp01} & $7.5$K & Teen & Student, Spy & United States & Comedy & \enquote{Look, I love you! I have loved you since the moment I saw you. Please! I'll never get a chance to tell you.} \\ \midrule
\textbf{sp02} & $3.9$K & Young Adult & Unemployed, Community Service & United Kingdom & Comedy, Drama & \enquote{In the words of the great Lionel Richie...hello.} \\ \midrule
\textbf{sp03} & $3.2$K & Adult & Actor & United States & Docuseries & \enquote{So be present, be spontaneous. Enjoy the moment, enjoy yourself and learn.} \\ \midrule
\textbf{sp04} & $3.1$K & Adult & Criminal Profiler & United States & Crime, Drama, Horror & \enquote{It isn't very smart to piss off a guy who thinks about killing people for a living.} \\ \midrule
\textbf{sp05} & $3.1$K & Adult & Psychiatrist & United States & Crime, Drama, Horror & \enquote{Before we begin, I must warn you... nothing here is vegetarian.} \\ \bottomrule
\end{tabular}}
\caption{Selected metadata regarding long-term speakers from \textsc{ZOO} used in the experiment.}
\label{tab:zoo-speakers}
\end{table*}

\begin{table*}[h!]
\centering
\resizebox{\linewidth}{!}{
\begin{tabular}{rp{6cm}lp{7cm}}
\toprule
\textit{\textbf{ID}} & \textit{\textbf{Top-gaining sentence (4+ words)}} & \textit{\textbf{Five top-gaining tokens}} & \textit{\textbf{Top-losing sentence (4+ words)}} \\ \midrule
\textbf{sp01} & \enquote{Paranoid and can fit into small spaces.} & Okay, Wait, spy, Mom, mission & \enquote{To teach and to lend a guiding hand.} \\ \midrule
\textbf{sp02} & \enquote{Fucking nuns! Fucking shit!} & Fuck, Shit, Fucking, fucking, fuck & \enquote{English, Math and French.} \\ \midrule
\textbf{sp03} & \enquote{I love this car.} & Wow, ital, coffee, brain, b & \enquote{I'm not opposed to doing things to my teeth.}  \\ \midrule
\textbf{sp04} & \enquote{One missing kid's a boy.} & killer, Jack, kill, close, life & \enquote{She was a slim and delicate pig.} \\ \midrule
\textbf{sp05} & \enquote{Is your conscience clear?} & got, killer, Will, Jack, Ab & \enquote{Simpler times in boatyards with dad.}  \\ \bottomrule
\end{tabular}}
\caption{Sentences and tokens for which the log likelihood under \textsc{LMCue} ($\mathcal{S}+\mathcal{P}$) changes the most compared \textsc{Base-LM}.}
\label{tab:samples}
\end{table*}

\begin{table}[h!]
\centering
\resizebox{\linewidth}{!}{
\begin{tabular}{r|c|ccccc|c}
\toprule
& \multirow{2}{*}{\textsc{sMRR}$^\uparrow$} & \multicolumn{6}{c}{\textsc{ppl}$^\downarrow$} \\
& & \textbf{sp01} & \textbf{sp02} & \textbf{sp03} & \textbf{sp04} & \textbf{sp05} & \textbf{Mean} \\ \midrule
\textit{\textbf{baselines}} &  &  &  &  &  &  & \\
\textit{\textbf{non-context}} &  &  &  &  &  &  & \\
\textsc{Base-LM} & $0.2$ & $15.24$ & $17.39$ & $23.53$ & $18.64$ & $21.14$ & $19.19$ \\
\textit{\textbf{fine-tuning}} &  &  &  &  &  &  & \\
\textsc{SpFineTune} & $\mathbf{1.0}$ & $14.54$ & $\mathbf{16.01}$ & $21.76$ & $17.36$ & $\underline{19.50}$ & $\mathbf{17.84}$\\
\textsc{Lerp} & $\mathbf{1.0}$ & $\mathbf{14.35}$ & $16.31$ & $22.25$ & $17.66$ & $19.91$ & $18.10$ \\ \midrule
\textit{\textbf{proposed}} &  &  &  &  &  & &  \\
\textit{\textbf{metadata-based}} &  &  &  &  &  & &  \\
\textsc{LMCue} ($\mathcal{S}$) & $\mathbf{1.0}$ & $14.99$ & $16.75$ & $21.86$ & $17.54$ & $19.89$ & $18.21$ \\
\textsc{LMCue} ($\mathcal{P}$) & $0.8$ & $14.68$ & $17.17$ & $\underline{21.26}$ & $\underline{17.12}$ & $\mathbf{19.45}$ & $17.94$ \\
\textsc{LMCue} ($\mathcal{S}+\mathcal{P}$) & $\mathbf{1.0}$ & $14.77$ & $16.77$ & $\mathbf{21.22}$ & $\mathbf{17.10}$ & $\underline{19.47}$ & $\underline{17.87}$ \\ \bottomrule
\end{tabular}}
\caption{Results on the test set for long-term speakers. \underline{Underlined} results are on par with results in \textbf{bold}.}
\label{tab:res-long-term}
\end{table}

To determine the effectiveness of our context-based adaptation approach when compared to LMs fine-tuned on the available speaker dialogue, we focus our setup on five long-term (multi-episode TV) speakers with at least 3,000 lines of dialogue sources from the \textsc{ZOO} corpus (\autoref{tab:zoo-speakers}). For each speaker, we use 400 and 600 of these lines for validation and testing, respectively. Within this experiment, we use \textsc{SpFineTune} and \textsc{Lerp} as baselines. We obtain \textsc{SpFineTune} by fine-tuning the LM on all \textsc{ZOO} data initially (FT1), and then once more on speaker data alone. \textsc{Lerp} is obtained from a mean interpolation of the prediction probabilities from \textsc{SpFineTune} and FT1. We fine-tune an individual model instance for each speaker $\in \{{\textbf{sp01}, ..., \textbf{sp05}}\}$.

\textsc{LMCue} achieves results comparable to all speaker-fine-tuned models (\autoref{tab:res-long-term}). When using speaker metadata ($\mathcal{S}$), \textsc{LMCue} achieves \textsc{sMRR} of 1.0 just like fine-tuned models, suggesting the perplexity improvements come from the model's context-based predictions. \textsc{LMCue} ($\mathcal{P}$) achieves lower \textsc{sMRR} (0.8): its predictions are based only on production metadata, not considering that two different characters may come from the same production. Any adapted model, whether fine-tuned or metadata-based, yields a reduction in perplexity between 5.1\% and 6.8\% which is comparable to results on \texttt{test}. \textsc{SpFineTune} achieves the best overall perplexity reduction of 1.32 and 1.0 \textsc{sMRR}, with \textsc{LMCue} ($\mathcal{S}+\mathcal{P}$) yielding a statistically comparable reduction of 1.29 and the same \textsc{sMRR} while requiring (i) no fine-tuning and (ii) the maintenance of only one model for all speakers.

To illustrate how personalisation manifests in practice, we identify the predictions of \textsc{LMCue} ($\mathcal{S}+\mathcal{P}$) with the most increased and decreased log likelihood compared to \textsc{Base-LM} (compare \autoref{tab:zoo-speakers} and \autoref{tab:samples}). Top-gaining tokens have strong associations with certain categories, like \textit{profession} (\textbf{sp01} \enquote{Student, Spy} $\rightarrow$ \textit{spy, Mom, mission}), \textit{age} (\textbf{sp02} \enquote{Young Adult} $\rightarrow$ expletives) or genre (\textbf{sp04, sp05} \enquote{Crime, Drama, Horror} $\rightarrow$ \textit{killer}). Similarly, top-gaining sentences for \textbf{sp01} and \textbf{sp02} have a comedic overtone (matching the genre), while the top-losing sentences do not fit these characters' demographic profiles.

For further analysis on this subject, \autoref{lrec:cost-benefit} shows the individual contribution of each character metadata type to reduction in perplexity.

\paragraph{Zero-shot Transfer} \label{lrec:rq2}
We assess the effectiveness of speaker adaptation for completely \textbf{new test speakers} featured in the \texttt{test\_unseen} sets of both corpora. To reiterate, these speakers' dialogue is excluded from training and validation data (although there are overlaps in production metadata). As before, we fine-tune the pre-trained \textsc{LMCue} on the \texttt{train} splits. We compare the performance only to \textsc{Base-LM} since other baselines are not equipped to work well in this zero-shot scenario.

\autoref{tab:results-unseen} shows that \textsc{LMCue} ($\mathcal{S}$) still improves perplexity over a parameter-matched LM. Though these improvements are smaller than in the supervised scenario, they are still significant, especially for \textsc{ZOO} ($-$0.62). More importantly, for both corpora $\mathcal{S}$ is strongly beneficial towards high speaker separation (i.e. the model assigns the highest probability to dialogue which matches the given speaker's profile), as measured by \textsc{sMRR}. Perplexity does improve more when $\mathcal{P}$ is also used (1.4 $\rightarrow$ 5.6\% for \textsc{Cornell-rich}, 3.6 $\rightarrow$ 4.4\% for \textsc{ZOO}), though in this scenario we are evaluating the easier task of modelling new speakers in seen or unseen productions. $\mathcal{P}$roduction metadata alone yields the best reduction of 6.9/5.3\%. Using it has a different effect on the two test sets: in \textsc{Cornell-rich} it induces a stronger boost in \textsc{sMRR} than $\mathcal{S}$ ($+$0.08), while in \textsc{ZOO} it decreases it considerably ($-$0.20); second, using it in conjunction with $\mathcal{S}$ results in best \textsc{sMRR} in \textsc{Cornell} (0.94), but not so for \textsc{ZOO}. This can be explained by the fact that \textsc{ZOO} uses a pool of only nine productions (vs 595 in \textsc{Cornell-rich}), so adding $\mathcal{P}$ on top of $\mathcal{S}$ is unlikely to increase speaker separation. In contrast, \textsc{Cornell-rich} uses a rich pool of films, so film metadata is more likely to be unique between any two speakers, thus introducing it separates the two speakers even more, increasing \textsc{sMRR}. This is also why \textsc{sMRR} is so high for \textsc{LMCue} ($\mathcal{P}$): with 24 unique films between the 30 speakers the film metadata is rarely shared between any two speakers, making their context inputs more dissimilar. The magnitude of improvements in \textsc{sMRR} is also different for the two corpora, which again could be attributed to scale (863 vs 159 speakers, 595 vs 9 productions). Increasing the number of annotated entities can therefore improve the personalisation effect. Nevertheless, a score of 0.39 still suggests that \textsc{LMCue} ranks an unseen character on the 2.56th position with a model built from their demographic profile, on average.

\begin{table}[t]
\centering
\resizebox{\linewidth}{!}{
\begin{tabular}{rcccc}
\toprule
 & \multicolumn{2}{c}{\textsc{Cornell-rich}} & \multicolumn{2}{c}{\textsc{ZOO}} \\
 & \multicolumn{2}{c}{\texttt{test\_unseen}} & \multicolumn{2}{c}{\texttt{test\_unseen}} \\ \midrule
Approach & \textsc{ppl}$^\downarrow$ & \textsc{sMRR}$^\uparrow$ & \textsc{ppl}$^\downarrow$ & \textsc{sMRR}$^\uparrow$ \\ \midrule
\textit{\textbf{baselines}} & & & & \\
\textsc{Base-LM} & $23.62$ & $0.03$ & $17.11$ & $0.09$ \\ \midrule
\textit{\textbf{proposed}} & & & & \\
\textsc{LMCue} ($\mathcal{S}$) & $23.28$ & $0.70$ & $16.49$ & $\mathbf{0.39}$ \\
\textsc{LMCue} ($\mathcal{P}$) & $\mathbf{22.00}$ & $0.80$ & $\mathbf{16.21}$ & $0.19$ \\
\textsc{LMCue} ($\mathcal{S} + \mathcal{P}$) & $22.31$ & $\mathbf{0.96}$ & $16.35$ & $0.32$ \\
 \bottomrule
\end{tabular}}
\caption{Results of evaluation with speaker \& film metadata on the test set of unseen speakers.}
\label{tab:results-unseen}
\end{table}

Using \textsc{LMCue} ($\mathcal{S}+\mathcal{P}$), we queried the words for which log likelihood increased the most w.r.t. \textsc{Base-LM} in the \texttt{test} set of \textsc{Cornell-rich} and obtained a list of the following fifteen tokens:
\begin{center}
	\textit{crew shark ship azz birds casino space leads \\power ocean camp boat cops baby ace}
\end{center}
Many of these tokens are context-specific and would only appear in certain scenarios or domains. For example, \textit{casino} or \textit{space} are unlikely to appear in a sentence unless they represent locations within the film. A subset of the provided tokens (\textit{crew, shark, ship, ocean, birds}) may also collectively describe a single scenario, such as an adventure or thriller film set on a ship in the middle of an ocean. We hypothesise that a few such films appeared in the training set \textsc{Cornell-rich}, allowing \textsc{LMCue} to develop a strong prior for predicting these tokens when metadata of similar films is provided as input. Finally, these tokens are notably more generic than those in \autoref{tab:res-long-term}: we observe that the effect of biasing speaker-specific vocabulary may be limited for some tokens compared to the supervised scenario (e.g. tokens representing names of the character's co-stars are not related to demographic features so would not be affected in a zero-shot scenario).

\section{Measuring Personalisation in Translation} \label{lrec:translations}
In \S \ref{lrec:lmcue} we have established empirically that \textsc{LMCue} exhibits effects of context-based personalisation, acting as a person- and production-specific LM when provided with their metadata, and is comparable with speaker-specific fine-tuning approaches. Compared to a general LM, it assigns higher probability to tokens which are more likely to occur in the given character and production context. Within this section, we use this model as a \enquote{contextual oracle}, applying it to various streams of dialogue to obtain judgements on how likely the dialogue is to be said in the given context. We also use a non-contextual LM as a \enquote{non-contextual oracle}, to measure the extent to which the given text co-occurs specifically with the provided context. We are interested in the following research question:
\begin{enumerate}[rightmargin=1em]
  \setlength{\itemindent}{2em}
    \item [\textit{\textbf{RQC}}] \textit{Can MT offer personalisation benefits proportional to professional translations?}
\end{enumerate}

We operate on four iterations of the same text: original version in English (\textsc{Original}), professional translations of the original text to French, German or Polish (\textsc{Reference}), and several versions of machine-translated text, which we describe below. Our goal is to establish to what extent the effect of personalisation (context-specificity to particular character and production descriptions) is found in professional and machine translations, and whether hypotheses generated by a contextual machine translation system exhibit stronger personalisation effects compared to non-contextual.

\subsection{Datasets}
\paragraph{The \textsc{ZOO-Multi} Corpus} \label{lrec:ZOO-multi}
We perform the experiments on \textsc{ZOO-Multi}, which is a subset of the episodes featured in the \textsc{ZOO} corpus, but which includes professional translations to French, German and Polish. This corpus is split differently into training, validation and testing subsets than \textsc{ZOO}. Specifically, we pre-select three TV series: an unscripted documentary, an unscripted cooking competition show and a scripted drama series and use between four and six episodes each for testing. Then we split the remaining data as follows, in two different data settings: \textsc{Disjoint}, where we use the remaining episodes as validation data, and all other TV series as training data (so that there is no overlap between training and validation/testing data) and \textsc{Overlap}, where we include the remaining episodes in the training data, and select random utterances from other series for validation. These two data settings represent the real-life scenarios of the subtitle translation task: when no past episodes are available for the considered series (\textsc{Disjoint}), and when there are some already completed translations that can be leveraged (\textsc{Overlap}) -- see \autoref{tab:data-quants2}, rows 1-8 for quantitative details and \autoref{app:zoo-multi-ex} for an individual sample.

\begin{table}[h!]
\centering
\resizebox{\linewidth}{!}{
\begin{tabular}{@{}clccc}
\toprule
& & \multicolumn{3}{c}{Number of samples} \\
Row & Dataset \& split & \textsc{en-fr} & \textsc{en-de} & \textsc{en-pl}  \\
\midrule
(1) & \textsc{ZOO-Multi} & & & \\
(2) & \hspace{1em}\textsc{Disjoint} & & & \\
(3) & \hspace{2em}\texttt{train}* & $58.5$K & $59.0$K & $107.1$K \\
(4) & \hspace{2em}\texttt{valid}* & $4.0$K 	& $3.8$K & $4.1$K \\
(5) & \hspace{1em}\textsc{Overlap} & & & \\
(6) & \hspace{2em}\texttt{train}* & $60.3$K & $60.8$K & $106.1$K \\
(7) & \hspace{2em}\texttt{valid}* & $2.3$K 	& $2.3$K & $2.3$K \\
(8) & \hspace{1em}\texttt{test} & $7.8$K 	& $7.8$K & $7.6$K \\
(9) & \textsc{OpenSubtitles} & & &  \\
(10) & \hspace{1em}\texttt{train} & $14.7$M & $5.3$M & $12.4$M \\

\bottomrule
\end{tabular}}
\caption[Quantities of segments, tokens and unique metadata in the \textsc{OpenSubtitles} and \textsc{ZOO} datasets.]{Quantities of segments in \textsc{ZOO} and \textsc{OpenSubtitles}. *Values are averaged over dataset iterations generated for each of the three series.}
\label{tab:data-quants2}
\end{table}

\paragraph{\textsc{OpenSubtitles} Corpus}
For pre-training, we re-use the version of \textsc{OpenSubtitles} corpus described in \citet{vincent-etal-2023-mtcue}. The dataset comprises sentence pairs annotated with six production metadata (via OMDb) and document-level data. Data quantities are listed in \autoref{tab:data-quants2}, row 10.
\subsection{Evaluation and Systems}
We evaluate how well adapted to context individual version of the text are by finding out the degree of co-occurrence between individual sentences or translations and their specific extra-textual context. We express results as the pointwise mutual information (PMI) between the context $\mathcal{C}$ and the target utterance or hypothesis $\mathcal{H}$, which is computed as:
\begin{align} 
    \text{PMI}(\mathcal{C}, \mathcal{H}) &= \log \frac{p(\mathcal{H} \mid \mathcal{C})}{p(\mathcal{H})} \\
    &= \log p(\mathcal{H} \mid \mathcal{C}) - \log p(\mathcal{H}) \label{eq:pmi} 
\end{align}

\noindent{PMI rewards positively those tokens which occur more frequently in the context $\mathcal{C}$ than in the general distribution termed with the prior $p(\mathcal{H})$. In practice, both terms of \autoref{eq:pmi} are computed with LMs: $\log p(\mathcal{H} \mid \mathcal{C})$ with \textsc{LMCue} ($\mathcal{S}+\mathcal{P}$) and $\log p(\mathcal{H})$ with \textsc{Base-LM}. Both LMs are pre-trained as described in \S \ref{lrec:lm-exp} and fine-tuned on the context and target-side dialogue from \textsc{ZOO-Multi} corpus (\textsc{Overlap} setting). We train a separate tandem of LMs for each language pair, and for statistical significance we train five distinct instances of each model (each with a different random seed). } 

We use the open source implementation of \textsc{MTCue} \citep{vincent-etal-2023-mtcue} to generate non-contextual and contextual machine translations. 
We also follow the authors’ implementation of a parameter-matched baseline (\textsc{Base-NMT}). We pre-train the systems on \textsc{OpenSubtitles} with metadata and document-level information (as per the original paper) 
and fine-tune them on the \textsc{ZOO-Multi} corpus for each language pair separately. We include training details in \autoref{app:lmcue-training}.

\subsection{Results} \label{lrec:rq4} \label{lrec:rq5}
A positive value of PMI for \textsc{Reference} (mean score of 0.073; \autoref{lrec:pmi}) suggests the presence of a co-occurrence effect between professional translations of the test set and their extra-textual context. We calculated that the \textsc{Original} text in English obtained a score of 0.087. The two values are not directly comparable (as computed on different language variants of the test set), but they are of similar magnitude which suggests that the context-specific traits of the original text are well preserved in the professional translations for this test set.

\begin{table}[h!]
\centering
\resizebox{\linewidth}{!}{
\begin{tabular}{@{}lcccc@{}}
\toprule
 & French & German & Polish & Mean \\ \midrule
\textsc{Reference} \hspace{1em} & $0.101$ & $0.037$ & $0.081$ & $0.073$ \\ \midrule
\textsc{Disjoint} & & & & \\
\hspace{2em} \textsc{Base-NMT} & $0.042$ & $-0.004$ & $0.041$ & $0.026$ \\
\hspace{2em} \textsc{MTCue} & $0.066$ & $0.007$ & $0.049$ & $0.041$ \\ \midrule
\textsc{Overlap} & & & & \\
\hspace{2em} \textsc{Base-NMT} & $0.040$ & $-0.006$ & $0.049$ & $0.028$ \\
\hspace{2em} \textsc{MTCue} & $\mathbf{0.069}$ & $\mathbf{0.012}$ & $\mathbf{0.072}$ & $\mathbf{0.051}$ \\ \bottomrule
\end{tabular}}
\caption{PMI computed with general and personalised LMs on translations. Results computed from five different runs.}
\label{lrec:pmi}
\end{table}

\begin{table}[h!]
\centering
\resizebox{\linewidth}{!}{
\begin{tabular}{@{}lcccccc@{}}
\toprule
 & \multicolumn{2}{c}{French} & \multicolumn{2}{c}{German} & \multicolumn{2}{c}{Polish} \\
 & \textsc{BLEU} & \textsc{Comet} & \textsc{BLEU} & \textsc{Comet} & \textsc{BLEU} & \textsc{Comet} \\ \midrule
\textsc{Disjoint} & & & & & & \\
\hspace{.5em} \textsc{Base-NMT} & $34.89$ & $23.69$ & $35.71$ & $28.59$ & $31.13$ & $31.50$ \\
\hspace{.5em} \textsc{MTCue} & $35.73$ & $25.81$ & $36.22$ & $29.29$ & $31.62$ & $32.66$ \\ \midrule
\textsc{Overlap} & & & & & & \\
\hspace{.5em} \textsc{Base-NMT} & $35.06$ & $23.75$ & $36.15$ & $29.75$ & $31.59$ & $\mathbf{32.83}$ \\
\hspace{.5em} \textsc{MTCue} & $\mathbf{36.14}$ & $\mathbf{27.04}$ & $\mathbf{36.90}$ & $\mathbf{30.53}$ & $\mathbf{32.18}$ & $31.95$ \\ \bottomrule
\end{tabular}}
\caption{\textsc{BLEU} and \textsc{Comet} scores for the evaluated MT systems. \textsc{Comet} score was computed using the \texttt{lrec20-comet-da} model.}
\label{lrec:bleu-comet}
\end{table}

\textsc{Base-NMT} achieves a significantly lower PMI on average (\autoref{lrec:pmi}, 0.026 for \textsc{Disjoint}, 0.028 for \textsc{Overlap}): the absence of context at generation time results in translations less adapted to the specific characters and productions. The nevertheless positive values can be explained by the presence of domain-specific terms such as \enquote{pan} which do not need context to be translated correctly but will occur more often in specific contexts (e.g. cooking shows), yielding a positive PMI. The contextual \textsc{MTCue} achieves higher PMI on average than \textsc{Base-NMT} ($+$0.015 for \textsc{Disjoint}, $+$0.023 for \textsc{Overlap}), meaning that using the relevant context does make the hypotheses more personalised, and the greater improvement for \textsc{Overlap} suggests that context can be utilised even better when previous samples from the same speakers or series are given. Interestingly, while personalisation is stronger in \textsc{Overlap} systems than \textsc{Disjoint}, the \textsc{MTCue} (\textsc{Disjoint}) still performs better than either \textsc{Base-NMT} system, signifying the robustness of \textsc{cue} vectors to this zero-shot adaptation to new series and speakers. Among all three target languages, reference translations to German are correlated with context the least (PMI $=$ 0.037) and translations to French the most (PMI $=$ 0.101). The MT systems' results follow a similar trend.
We also provide the \textsc{BLEU} and \textsc{Comet} scores, comparing the MT to the human references (\autoref{lrec:bleu-comet}): in both data settings, \textsc{MTCue} matches the references to a significantly higher extent than the baseline. 

\section{Conclusions} \label{lrec:conclusions}
We have argued for context-based personalisation of LMs by training a conditional generation architecture on dialogue accompanied by rich contextual annotations. We have also explored using such models to evaluate the context-specificity of professional and machine translations, providing insight into how well the generated translations are specific to the extra-textual context, without direct comparison to the human references. Finally, we have contributed \textsc{Cornell-rich}, a set of rich speaker and production annotations for a publicly available dialogue dataset. Below we summarise the findings specific to each research question.
\vspace{-.2em}
\begin{enumerate}[rightmargin=1em]
  \setlength{\itemindent}{2em}
    \item [\textit{\textbf{RQA}}] \textit{How can rich character profiles be used to model the characters' speaking styles?}
\end{enumerate}
\vspace{-.2em}
The \textsc{LMCue} architecture can be trained to exhibit personalisation leveraging the speaker and production metadata. This approach is effective in a few-shot scenario, matching speaker-specific fine-tuning performance, while requiring no such procedure. This makes it the more effective method if metadata is available.
\vspace{-.2em}
\begin{enumerate}[rightmargin=1em]
  \setlength{\itemindent}{2em}
    \item [\textit{\textbf{RQB}}] \textit{How can a LM be personalised for a specific character solely by learning from data for characters with similar profiles?}
\end{enumerate}
\vspace{-.2em}
On a test set with unseen speakers, context-based personalisation yields a high \textsc{sMRR} score, (i.e. speaker separation). When both speaker and production metadata is used, the models achieve a reduction in perplexity comparable with the seen speaker scenario. Therefore, unlike speaker-specific fine-tuning, our method is applicable to new characters. We also found that having a varied pool of speakers and productions in training data correlates positively with \textsc{sMRR}.
\vspace{-.2em}
\begin{enumerate}[rightmargin=1em]
  \setlength{\itemindent}{2em}
    \item [\textit{\textbf{RQB}}] \textit{Can MT offer personalisation benefits proportional to professional translations?}
\end{enumerate}
\vspace{-.2em}
Utilising speaker and metadata annotations in MT makes the language used in hypotheses more context-specific, as measured by the PMI score between such context and the generated text, when compared to a context-agnostic system. However, this context specificity is still stronger in gold standard (professional) translations. Our findings suggest that contextual LMs could be paired with automatic metrics for a more well-rounded evaluation of machine translation as they bring the aspect of the translations fitting the specific extra-textual context.

\section{Bibliographical References}\label{sec:reference}
\bibliographystyle{lrec-coling2024}
\bibliography{bibliography/custom,bibliography/anthology,bibliography/references}

\appendix
\section{Details regarding the data collection campaign} \label{app:data-collection}
The data collection process was carried out by two annotators, both native English speakers and experts in the dubbing and subtitling industry.
After parsing \textsc{Cornell}\footnote{\url{https://convokit.cornell.edu/documentation/movie.html}}, a spreadsheet of characters was generated that included their name, source film, and the number of lines attributed.

From previous work \citep[e.g.][]{johannsen-etal-2015-cross} and hypotheses made based on experts' experience, we pre-defined a number of categories of information to collect about each character. Specifically, we selected categories that we hypothesised to (i) be identifiable from the available sources and (ii) influence a person's speaking style or vocabulary used. They were: their \textbf{age bracket} $\in \{\text{child, teen, young adult, adult, elderly}\}$, \textbf{profession}, \textbf{character description} (a few sentences summarising their personality or character arc), \textbf{religion} and a \textbf{characteristic quote}: a typical or quotable thing the character might say. Additionally, the \textbf{gender} annotations from the original corpus were re-used, and an optional column \enquote{\textbf{additional information}} was included to collect comments from experts\footnote{Upon inspection: the annotators predominantly used this field to provide the actor's name, an interesting fact about the character (e.g. \enquote{Plays a caricature of himself}), or trivia.}. The characters with the most lines spoken were prioritised, resulting in all 714 characters with 100+ lines and 149 with 50$-$99 lines being annotated, totalling 863 characters.

\paragraph{Annotation sources} Annotations are based on publicly available pages from Wikipedia\footnote{\url{https://wikipedia.org/}} for individual films, as well as fan-made Fandom\footnote{\url{https://fandom.com/}} pages for both films and characters. Where information was unavailable from these sources, the annotators either referred back to the corpus itself or skipped the given field altogether. The film metadata was obtained via the OMDb API.\footnote{\url{https://omdbapi.com/}}

\paragraph{Annotation decisions} The annotation process involved matching every script's name against an IMDb entry, which did not always yield a match as some scripts had been scrapped or rewritten or characters' names had been changed. Unidentifiable films and characters were not considered for annotation.
Some information, especially \textit{religion}, was occasionally difficult to find, in which case it would be skipped or labelled as \textit{Unknown}. It was challenging to produce annotations for characters based on real people, or for a real person played by themselves. Where characters were based on historical figures, the annotators focused on the production interpretation of the person; when dealing with a characterisation of the person at a specific point in time, the focus was on their behaviour at that point in time. Finally, some characteristics were unsuitable for selected character information: e.g. when a character was immortal, it did not fit into set age brackets, and for some characters there were limited clues to determine their age bracket. In both cases, the final annotations were based on the annotators' expertise.

\section{Details regarding the \textsc{Cornell-rich} Corpus} \label{app:cornell-details}
\textsc{Cornell-rich} comprises annotations for 863 speakers (speaker \textit{profiles}), covering 135.7K utterances; nearly half of the annotated speakers have 150$+$ lines of dialogue and about 25\% have 200$+$. At least 64.1\% of conversational exchanges feature at least one annotated character and as much as 95.5\% of the featured films are annotated with film metadata (\autoref{tab:cornell-speaker-details}). We provide a full list of collected metadata with examples in \autoref{tab:examples}.

\begin{table}[h!]
\centering
\resizebox{\linewidth}{!}{
\begin{tabular}{@{}r|rc|cc@{}}
\toprule
& Count & \% of all & Utterances & \% of all \\ \midrule
\midrule
\textbf{speakers} & $9.0$K & $-$ & $304.3$K & $-$ \\
\annedspeaker & $863$ & $9.5\%$ & $135.7$K & $44.6\%$ \\
\bustspeaker & $8.2$K & $90.5\%$ & $170.1$K & $55.9\%$ \\ \midrule
\textbf{exchanges} & $83.1$K & $-$ & $304.3$K & $-$ \\
\annedspeaker $\Leftrightarrow$ (\bustspeaker or \annedspeaker) & $53.3$K & $64.1\%$ & $202.4$K & $66.5\%$ \\
\bustspeaker $\Leftrightarrow$ \annedspeaker & $36.8$K & $44.3\%$ & $134.4$K & $44.2$\% \\
\annedspeaker $\Leftrightarrow$ \annedspeaker & $16.5$K & $19.8\%$ & $68.0$K & $22.3\%$ \\ \midrule
\midrule
\textbf{films} & $617$ & $-$ & $304.3$K & $-$ \\
annotated & $589$ & $95.5\%$ & $291.0$K & $95.6\%$ \\ \bottomrule
\end{tabular}}
\caption[Details of annotations compared to data quantities from \textsc{Cornell}.]{Details of annotations compared to data quantities from \textsc{Cornell}. \annedspeaker $=$ speaker with rich annotations. \bustspeaker $=$ speaker without rich annotations.}
\label{tab:cornell-speaker-details}
\end{table}

\begin{table}[h!]
\centering
\resizebox{\linewidth}{!}{
\begin{tabular}{@{}rp{7cm}@{}}
\toprule
Metadata type & Value \\ \midrule
\textbf{Speaker metadata} & \\
Gender & A man \\
Age bracket & Adult \\
Profession & Attorney \\
Description & Galvin graduated from Boston College's law school. Galvin had a promising legal career ahead of him at an elite Boston law firm until he was framed for jury tampering by a partner due to his plans to expose the firm's underhanded activities. (...) \\
Quote & Your honor, with all due respect: if you're going to try my case for me, I wish you wouldn't lose it. \\
Country of origin & USA \\
Religion & Christian \\
\textbf{Film metadata} & \\
Genre & Comedy, Drama \\
PG Rating & PG Rating: R \\
Names of writers & Written by: Paul Andréota, André Cayatte, Henri Coupon \\
Country of production & France, Italy \\
Year of release & Released in 1974 \\
Plot description & A French judge try to acquit a man who is accused of murdering his lover. \\ \bottomrule
\end{tabular}}
\caption{A sample from \textsc{Cornell-rich} with each type of collected metadata.}
\label{tab:examples}
\end{table}

\begin{figure*}[t]
  \centering
 \resizebox{\linewidth}{!}{
  \subfloat[Number of lines per production year of films in the corpus.]
  {
	\includegraphics[width=0.3\linewidth]{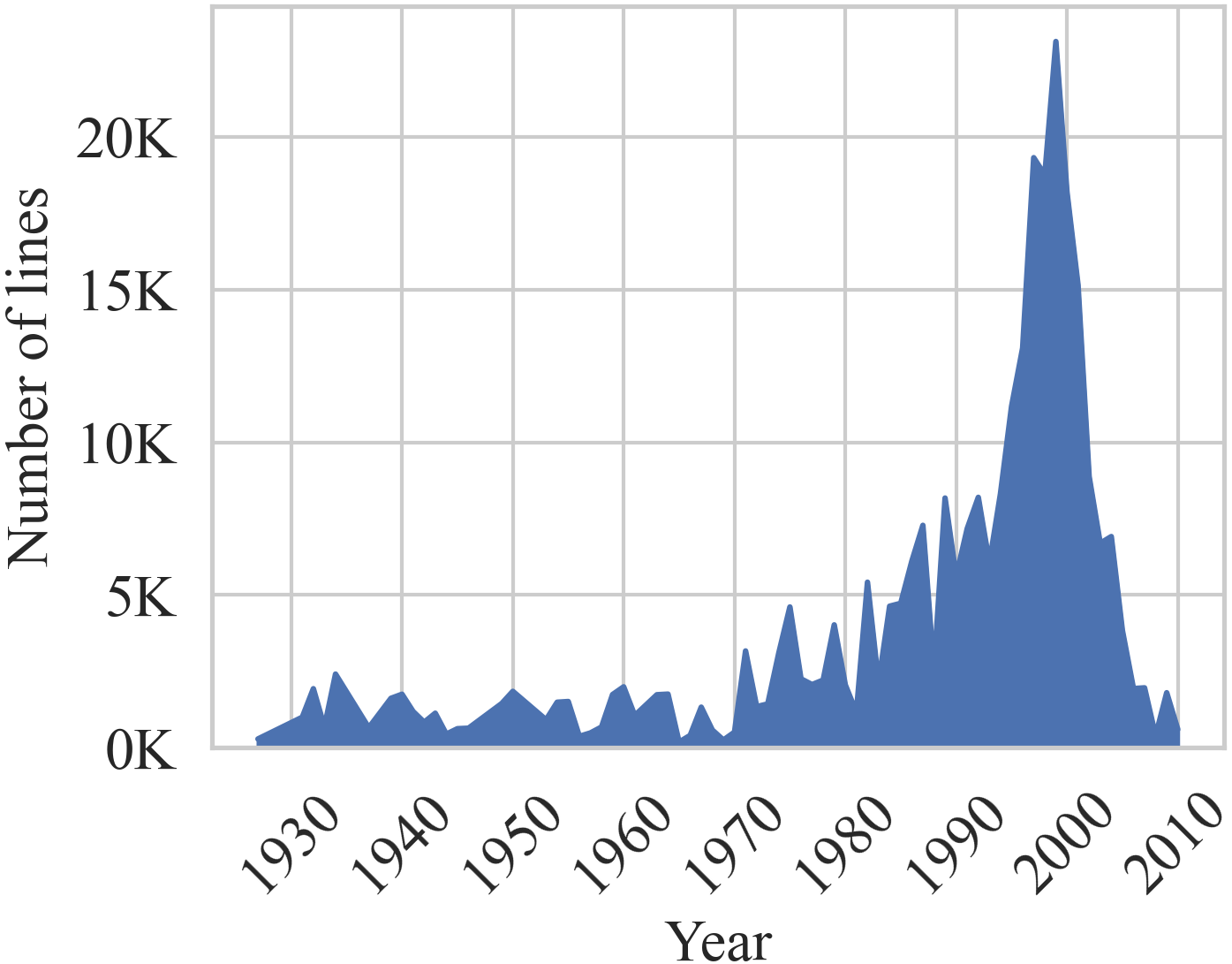}
   \label{lrec:years}
  }
  \hspace{0.3cm}
  \subfloat[15 most popular professions in the corpus.]
  {
	\includegraphics[width=0.32\linewidth]{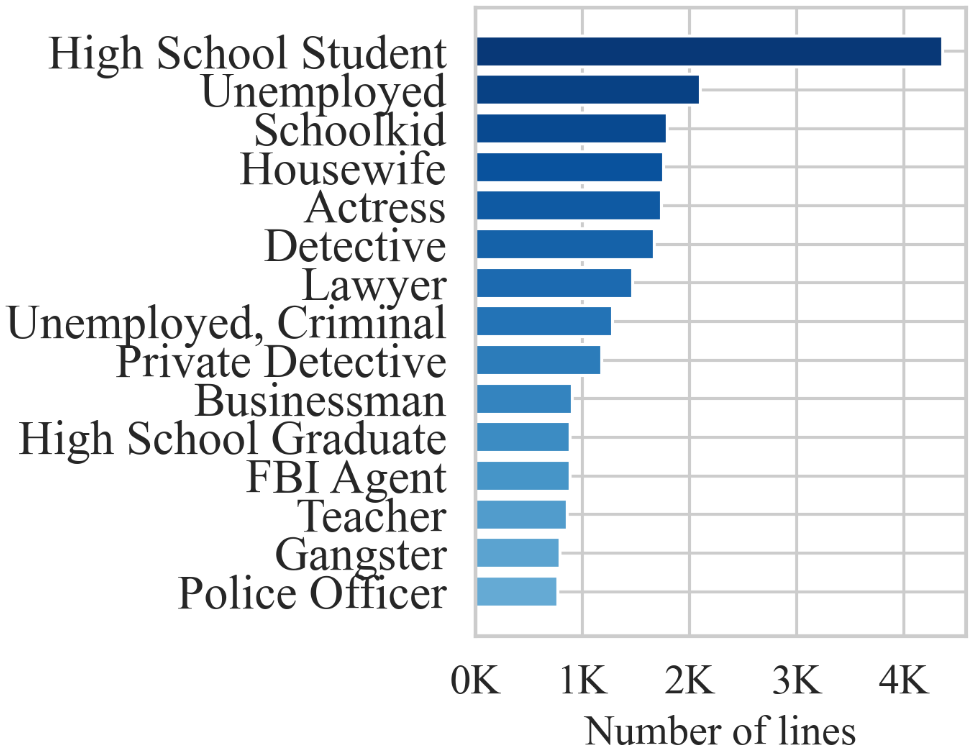}
	\label{lrec:profs}
  }
  \hspace{0.3cm}
  \subfloat[12 most popular genres in the corpus. Titles labelled as multiple genres are counted multiple times.]
  {
	\includegraphics[width=0.32\linewidth]{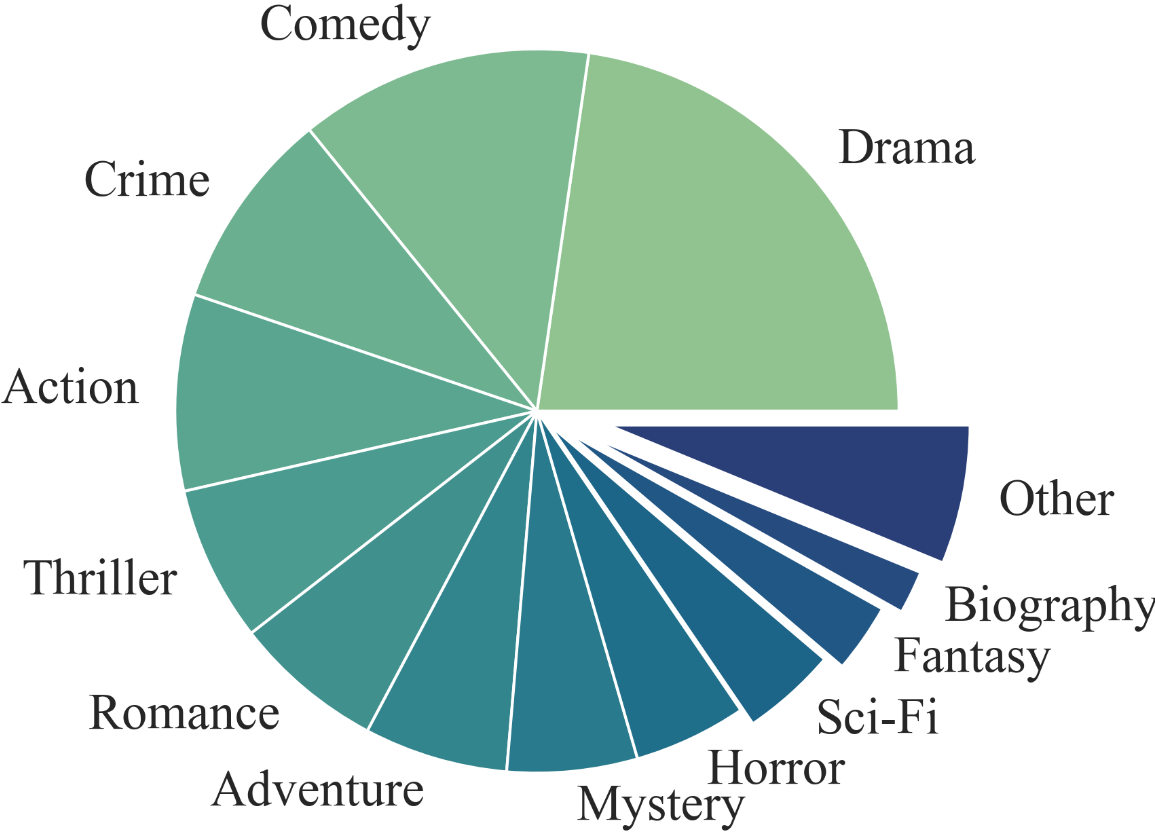}
	\label{lrec:genres}
  }}
  \caption{Visualisation of a subset of features of the proposed corpus.}
  \label{lrec:movie-details}
\end{figure*}

As per \autoref{lrec:movie-details}, the annotated films span nearly a century, with most lines coming from between the 1990s and 2005; the distribution of professions is significantly flatter, with the dominant field (\enquote{High School Student}) only making up about 3\% of the corpus. Finally, the most popular genres include drama, comedy, crime, and action.

\section{Pre-training Strategy: Past Dialogue as Proxy for Metadata} \label{app:pretraining}
This supplementary section presents empirical evidence that past dialogue can be used as a proxy for fine-tuning \textsc{LMCue} on speaker or production metadata. When fine-tuning, we use both $\mathcal{S}$peaker and $\mathcal{P}$roduction metadata. We report performance on \texttt{test\_unseen} to also present \textsc{sMRR} scores.

\begin{table}[h!]
\centering
\resizebox{\linewidth}{!}{
\begin{tabular}{r|cc|c|cc}
\toprule
 & \multicolumn{2}{c|}{Pre-training} & \multirow{2}{*}{Fine-tune} & \multirow{2}{*}{\textsc{ppl}$^\downarrow$} & \multirow{2}{*}{\textsc{sMRR}$^\uparrow$} \\
 & \xmark\hspace{.4em}/ \cmark & Context & & & \\ \midrule
 \textsc{Base-LM} & \cmark & $-$ & \xmark & $\mathbf{28.78}$ & $0.03$ \\
\textsc{LMCue} & \cmark & dialogue & \xmark & $37.19$ & $0.29$ \\
 \textsc{LMCue} & \cmark & metadata & \xmark & $30.95$ & $\mathbf{0.43}$ \\ \midrule
\textsc{Base-LM} & \xmark & $-$ & \cmark & $\mathbf{39.60}$ & $0.03$ \\
\textsc{LMCue} & \xmark & $-$ & \cmark & $51.14$ & $0.03$ \\ \midrule
\textsc{Base-LM} & \cmark & $-$ & \cmark & $23.62$ & $0.03$ \\
\textsc{LMCue} & \cmark & dialogue & \cmark & $\mathbf{22.31}$ & $\mathbf{0.96}$ \\
 \textsc{LMCue} & \cmark & metadata & \cmark & $22.71$ & $0.89$ \\ \bottomrule
\end{tabular}}
\caption{Results on \texttt{test\_unseen} of \textsc{Cornell-rich} from different pre-training/fine-tuning setups. New results (top 5 rows) come from single runs.}
\label{tab:pre-training}
\end{table}

As per \autoref{tab:pre-training}, pre-training on \textsc{OpenSubtitles} leads to best perplexity when no context is used (\textsc{Base-LM}), however using context yields improvements in \textsc{sMRR}, and these are stronger when metadata is used instead of dialogue. Similarly, without pre-training we also obtain the best perplexity with \textsc{Base-LM}; here even \textsc{sMRR} remains at a baseline level, i.e. the contextual model fails to learn contextual dependencies correctly. Metadata only leads to superior results when both pre-training and fine-tuning are included in the pipeline. Interestingly, 
pre-training on dialogue yielded the best results, though pre-training on metadata is not far behind ($+$0.4 \textsc{ppl}, $-$0.07 \textsc{sMRR}). We hypothesise that since past dialogue is much more diverse than film metadata (which contains many repeated fields), it is overall the better pre-training proxy for fine-tuning on new types of metadata, such as speaker profiles. For applications on other datasets, we therefore recommend pre-training on a similar dataset (domain-wise) with access to document-level information.
\section{Training details} \label{app:lmcue-training}
\subsection{\textsc{LMCue} and \textsc{Base-LM}}
We use off-the-shelf model architectures with pre-defined hyperparameters in \textsc{Fairseq} and only tune on three values each for batch size (simulated 200K to 400K tokens) and learning rate (3$\mathrm{e}-$4 to 1$\mathrm{e}-$3) based on validation performance on \texttt{valid} in \textsc{Cornell-rich}. For fine-tuning, we separately adapt these parameters for each dataset and metadata combination: learning rate (5$\mathrm{e}-$5 to 1$\mathrm{e}-$3) and batch size (0.25K to 20K tokens). The best fine-tuning set of learning rate and batch size was 5$\mathrm{e}-$5 and 1.5K for \textsc{LMCue} and 2$\mathrm{e}-$3 and 3K for \textsc{Base-LM}. Each model was trained on a single 32GB V100 GPU with an early stopping condition of validation loss not improving for 5 epochs. Pre-training \textsc{LMCue} and \textsc{Base-LM} took 35 and 17.5 GPU hours respectively while fine-tuning these models took respectively 0.78 and 0.32 GPU hours on average.

\subsection{\textsc{MTCue} and \textsc{Base-NMT}}
The machine translation models were trained in a similar way to the language models. We use the dataset and follow the pre-training regimen described in \citet{vincent-etal-2023-mtcue}. For fine-tuning, we separately adapt these parameters for each dataset and metadata combination: learning rate (5$\mathrm{e}-$5 to 1$\mathrm{e}-$3) and batch size (1K to 80K tokens). The best fine-tuning set of learning rate and batch size was 1$\mathrm{e}-$3 and 80K for \textsc{MTCue} and 3$\mathrm{e}-$3 and 20K for \textsc{Base-NMT}. Each model was trained on a single 32GB V100 GPU with an early stopping condition of validation \textsc{bleu} not improving for 5 epochs.

\section{Cost-benefit Analysis of Human Annotations} \label{lrec:cost-benefit}
Granular manual annotations are costly to obtain. Cost-benefit analysis helps avoid the misallocation of limited annotation funding and resources. This section presents the results of the cost-benefit analysis we conducted to show which individual speaker attributes produce the most \textbf{benefit} (reduction in perplexity) w.r.t. the perceived \textbf{cost} of producing them.

\begin{figure*}[htb!]
    \centering
    \resizebox{.8\linewidth}{!}{
    \includegraphics{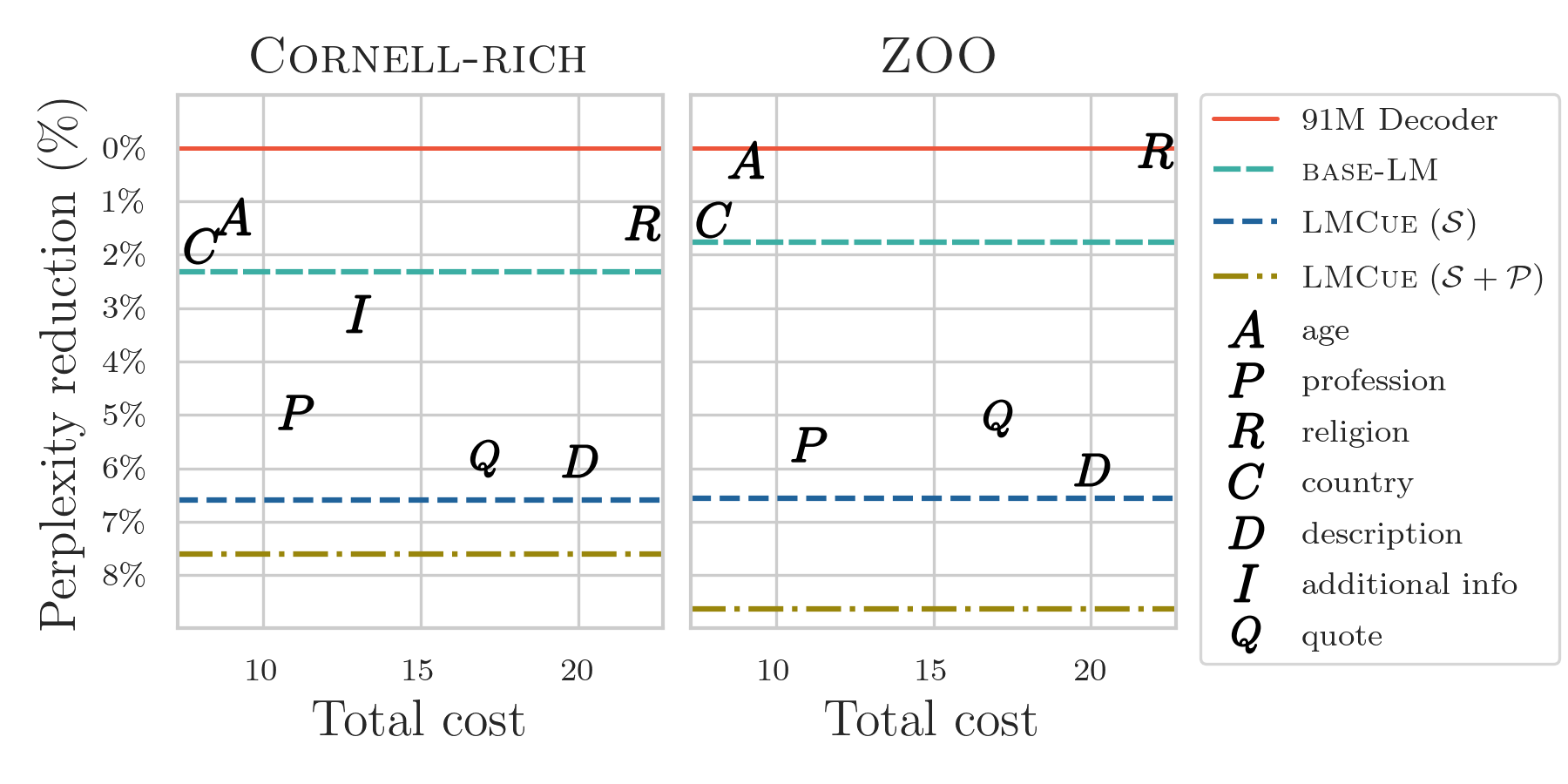}}
    \caption[Perplexity reduction from training LMCue with individual speaker attributes.]{Perplexity reduction from training \textsc{LMCue} with individual speaker attributes.}
    \label{lrec:cost-benefit-abl}
\end{figure*}

We asked the two human annotators to assess the effort required for the annotation task using three metrics on a Likert scale of 1 to 10: \textit{access} (how difficult it was to find information), \textit{credibility} (how confident they were in the accuracy/usefulness of the information), and \textit{time} (how much time was needed relative to other fields). We took the mean of both annotators' scores after reversing credibility ($C = 10 - C + 1$). We then conducted a simple experiment to measure the \textbf{benefit} of each metadata type by fine-tuning the pre-trained \textsc{LMCue} on each speaker metadata type evaluated individually. Finally, we measured the reduction in perplexity from including this information (\autoref{lrec:cost-benefit-abl}) compared to the $91$M parameter decoder in \textsc{LMCue}, since that is the decoder we are trying to improve with context.

The figure suggests that \textit{description}, \textit{profession} and \textit{quote} yield the greatest perplexity reduction in both datasets, around $5$ to $6\%$. \textit{Description}, the best-performing attribute, alone achieves $88.7/91.9\%$ of the perplexity reduction of \textsc{LMCue} $(\mathcal{S})$. On the other hand, \textit{age bracket}, \textit{religion} and \textit{country of origin} yield the smallest improvements, and a better improvement can be achieved with the parameter-matched \textsc{base-LM}. For \textsc{Cornell-rich}, they still help marginally ($1$ to $2\%$), while for \textsc{ZOO} improvements from \textit{age} and \textit{religion} are negligible. This analysis suggests why \citet{king-cook-2020-evaluating}, who implemented context-based adaptation using only \textit{age} and \textit{gender}, found it inferior to other methods; we found other variables such as \textit{description} to be significantly more useful.

Other findings of the analysis are consistent among the two corpora. \textit{Profession}, collected at a relatively small cost, is one of the top-3 attributes for both, hence the most cost-effective. \textit{Religion} is the least cost-effective attribute, requiring the most effort but producing the least benefit. Finally, experimental attributes such as characteristic \textit{quote}s and \textit{additional information}\footnote{Since \textit{additional information} was not collected for \textsc{ZOO}, it is not present on the \textsc{ZOO} plot of \autoref{lrec:cost-benefit-abl}.}
have been shown to be useful, the latter placing in the middle of the ranking whilst the former is on par with the best attribute for \textsc{Cornell-rich}. 

\section{Example from the \textsc{ZOO-Multi} corpus} \label{app:zoo-multi-ex}
In \autoref{tab:ZOO-Multi-stats} we show what a sample from the \textsc{ZOO-Multi} corpus looks like, including an example for each annotation as well as how many samples were annotated with the given context type.
\begin{table}[h!]
    \centering
    \resizebox{\linewidth}{!}{
    \begin{tabular}{r|cp{5cm}}
  \toprule
  Type of text field & Anno. \% & Example \\ \midrule
  English (source) & $-$ & That will take 12 hours. \\
  French translation & $-$ & Cela prendra 12 heures. \\ 
    German translation & $-$ & Es wird 12 Studen dauern. \\ 
  Polish translation & $-$ & To zajmie 12 godzin. \\ 
  \midrule
  \textbf{Production} & & \\
  country & $87.1\%$ & United States \\
  genre & $87.1\%$ & Drama, History \\
  PG rating & $36.4\%$ & PG rating: TV-14 \\
  plot description & $75.9\%$ & U.S. fighter pilots are recruited to test experimental aircraft and rockets to become first Mercury astronauts.   \\
  writers' names & $31.6\%$ & Written by: Mark Lafferty \\
  year & $87.1\%$ & Released in 2020 \\ \midrule
    \textbf{Speaker} & & \\
    age bracket & $50.3\%$ & Adult \\
    description & $50.3\%$ & Chris Kraft is an American aerospace and NASA engineer who was instrumental in establishing the agency's Mission Control Center and shaping its organization and culture. He is the Flight Director in charge of getting the mission control team ready for the launch of the ‘Mercury Seven’ team. \\
    characteristic quote & $39.5\%$ & He just kited a damn check. \\
    country of origin & $50.3\%$ & United States \\
    gender & $95.5\%$ & I am a man \\
    profession & $50.3\%$ & Flight Director \\
    religion & $13.5\%$ & Christian \\ \bottomrule
    \end{tabular}}
    \caption{Details regarding the \textsc{ZOO-Multi} corpus. Quantity refers to the number of samples within the training sets. Certain examples have been shortened for brevity.}
    \label{tab:ZOO-Multi-stats}
\end{table}

\section{Examples of Evaluation} \label{app:examples-eval}
In this section, we show some examples where the hypotheses from the contextual model (\textsc{MTCue}) were deemed more contextual (i.e. higher PMI) than the hypotheses from \textsc{Base-NMT}. All outputs come from systems in the \textsc{Overlap} setting and the English-to-Polish translation direction.

\begin{table}[h!]
    \centering
    \resizebox{\linewidth}{!}{
    \begin{tabular}{rp{0.9\linewidth}}
        \textbf{Example A} & \\ \midrule
        \textbf{Context}    & \textit{Family cooking competition show.} \\
        \textbf{Source}     & We're okay, we're doing just fine, just... \\[.5em]
        \textbf{Reference}  & Wszystko jest w porządku. \\
                            & \textit{('Everything is alright.')} \\ \midrule
        \textsc{\textbf{Base-NMT}}  & (\xmark) Nic nam nie jest. \textit{('We are fine.')} \\[.2em]
                                    & \hfill $-0.35$ PMI \hspace{1em} $-1.29$ \textsc{Comet} \\ \midrule
        \textsc{\textbf{MTCue}}     & (\cmark) Radzimy sobie. \textit{('We're coping.')} \\[.2em]
                                    & \hfill $\mathbf{0.32}$ PMI \hspace{1em} $\mathbf{-0.10}$ \textsc{Comet} \\
    \end{tabular}}
    \notag
\end{table}
\label{exa}

\hyperref[exa]{\textbf{Example A}} exhibits a behavioural agreement adaptation. During cooking show walkthroughs there is frequently talk of managing stress and working under pressure. It is done in good faith to let the contestants discuss their struggle. In Example 1, the non-contextual translation \textit{Nic nam nie jest.} would be far less likely to occur in this walkthrough setting than \textit{Radzimy sobie.} as this phrase is applicable to immediate impact situations (e.g. falls) rather than stressful, ongoing situations like a competition show. It also has a more defensive tone which is not a good fit for a lighthearthed family series.

\begin{table}[h!]
    \centering
    \resizebox{\linewidth}{!}{
    \begin{tabular}{rp{0.9\linewidth}}
        \textbf{Example B} & \\ \midrule
        \textbf{Context}    & \textit{Family cooking competition show.} \\
        \textbf{Source}     & Best think up some new recipes. \\[.5em]
        \textbf{Reference}  & Potrzebujemy nowych przepisów. \\
                            & \textit{('We need new recipes.')} \\ \midrule
        \textsc{\textbf{Base-NMT}}  & (\xmark) Lepiej wymyśl nowe przepisy. \\
                                    & \textit{('[You\textsubscript{s}] better think up some new recipes.')} \\[.2em]
                                    & \hfill $-0.76$ PMI \hspace{1em} $\mathbf{0.59}$ \textsc{Comet} \\ \midrule
        \textsc{\textbf{MTCue}}     & (\cmark) Wymyślmy nowe przepisy. \\
                                    & \textit{('Let's think up some new recipes.')} \\[.2em]
                                    & \hfill $\mathbf{-0.51}$ PMI \hspace{1em} $0.48$ \textsc{Comet} \\
    \end{tabular}}
    \notag
\end{table}
\label{exb}

In \hyperref[exb]{\textbf{Example B}}, \textsc{MTCue} guesses the number of addressees correctly: in this family cooking competition show, when someone is congratulated it is usually a family, i.e. a group of people. In our case-by-case analysis we continually found that the personalised LM exhibits a preference for correct gender and plurality.

\begin{table}[h!]
    \centering
    \resizebox{\linewidth}{!}{
    \begin{tabular}{rp{0.9\linewidth}}
        \textbf{Example C} & \\ \midrule
        \textbf{Context}    & \textit{Family cooking competition show.} \\
        \textbf{Source}     & Try the balls. \\[.5em]
        \textbf{Reference}  & Spróbuj kulkę. \textit{('Try a ball.')} \\ \midrule
        \textsc{\textbf{Base-NMT}}  & (\xmark) Spróbuj piłeczek. \textit{('Try the \textcolor{BrickRed}{footballs}.')} \\[.2em]
                                    & \hfill $-0.55$ PMI \hspace{1em} $\mathbf{0.51}$ \textsc{Comet} \\ \midrule
        \textsc{\textbf{MTCue}}     & (\xmark) Spróbuj jajek. \textit{('Try the \textcolor{BrickRed}{eggs}.')} \\[.2em]
                                    & \hfill $\mathbf{1.02}$ PMI \hspace{1em} $-0.19$ \textsc{Comet} \\
    \end{tabular}}
    \notag
\end{table}
\label{exc}

Finally, in \hyperref[exc]{\textbf{Example C}}, the language model assigned a significantly stronger PMI score to the hypothesis which translated \textit{balls} as \textit{eggs} instead of \textit{footballs}. While neither hypothesis is correct (the \enquote{balls} in the original text most likely referred to meatballs, or dough balls), the word \textit{eggs} is highly specific to the context of cooking, and therefore rated very highly by the personalised LM. This example shows that our method must be used in tandem with a standard translation quality metric such as \textsc{bleu} or \textsc{Comet}, as it is a monolingual evaluation tool that disregards the faithfulness to the source sentence.

\end{document}